\documentclass{article} 
\usepackage{colm2024_conference}

\usepackage{microtype}
\usepackage{hyperref}
\usepackage{url}
\usepackage{booktabs}
\usepackage{graphicx}
\definecolor{darkblue}{rgb}{0, 0, 0.5}
\hypersetup{colorlinks=true, citecolor=darkblue, linkcolor=darkblue, urlcolor=darkblue}

\usepackage{microtype}
\usepackage{algorithm}
\usepackage{algorithmic}
\usepackage[utf8]{inputenc} 
\usepackage[T1]{fontenc}    
\usepackage{url}            
\usepackage{booktabs}       
\usepackage{amsfonts}       
\usepackage{nicefrac}       
\usepackage{microtype}      
\usepackage{xcolor}         
\usepackage{enumitem}
\usepackage{multirow}
\usepackage{graphicx}
\usepackage{amsmath}

\usepackage{times}
\usepackage{latexsym}
\usepackage{multirow}
\usepackage{tabu}
\usepackage{booktabs}
\usepackage{graphicx}
\usepackage{amssymb}
\usepackage{pifont}
\newcommand{\cmark}{\ding{51}}%
\newcommand{\xmark}{\ding{55}}%
\usepackage{soul}
\usepackage{amssymb}

\title{\centering{Towards \textsc{LogiGLUE}:}\\ A Brief Survey and A Benchmark for Analyzing Logical Reasoning Capabilities of Language Models}

\author{Man Luo, Shrinidhi Kumbhar\thanks{ Equal Contribution}, Ming shen, Mihir Parmar, Neeraj Varshney, Chitta Baral \\
School of Computing and Augmented Intelligence\\
Arizona State University, Tempe, USA\\
\texttt{\{mluo26, skumbha4\}@asu.edu} \\
\And
Pratyay Banerjee \\
Amazon Alexa AI, USA \\
\AND
Somak Aditya \\
IIT KGP, India
}

\colmfinalcopy 
\begin{document}

\maketitle

\begin{abstract}
Logical reasoning is fundamental for humans yet presents a substantial challenge in the domain of Artificial Intelligence. Initially, researchers used Knowledge Representation and Reasoning (KR) systems that did not scale and required non-trivial manual effort.  Recently, the emergence of large language models (LLMs) has demonstrated the ability to overcome various limitations of formal Knowledge Representation (KR) systems. Consequently, there's a growing interest in using LLMs for logical reasoning via natural language. 
This work strives to understand the proficiency of LLMs in logical reasoning by offering a brief review of the latest progress in this area; with a focus on the logical reasoning datasets, tasks, and the methods adopted to utilize LLMs for reasoning. To offer a thorough analysis, we've compiled a benchmark titled \textsc{LogiGLUE}. This includes 24 varied datasets encompassing deductive, abductive, and inductive reasoning. 
Utilizing \textsc{LogiGLUE} as a foundation, we have trained an instruction fine-tuned language model, resulting in LogiT5. We study single-task training, multi-task training, and a ``chain-of-thought'' knowledge distillation fine-tuning technique to assess the model’s performance across the different logical reasoning categories. We also assess various large language models (LLMs) using LogiGLUE, and the findings indicate that LLMs excel most in abductive reasoning, followed by deductive reasoning, while they are least effective at inductive reasoning.
We aim to shed light on the capabilities and potential pathways for enhancing logical reasoning proficiency in LLMs, paving the way for more advanced and nuanced developments in this critical field.
\end{abstract}

\section{Introduction}

\begin{figure}[!htp]
    \centering
    \includegraphics[width=0.7\textwidth]{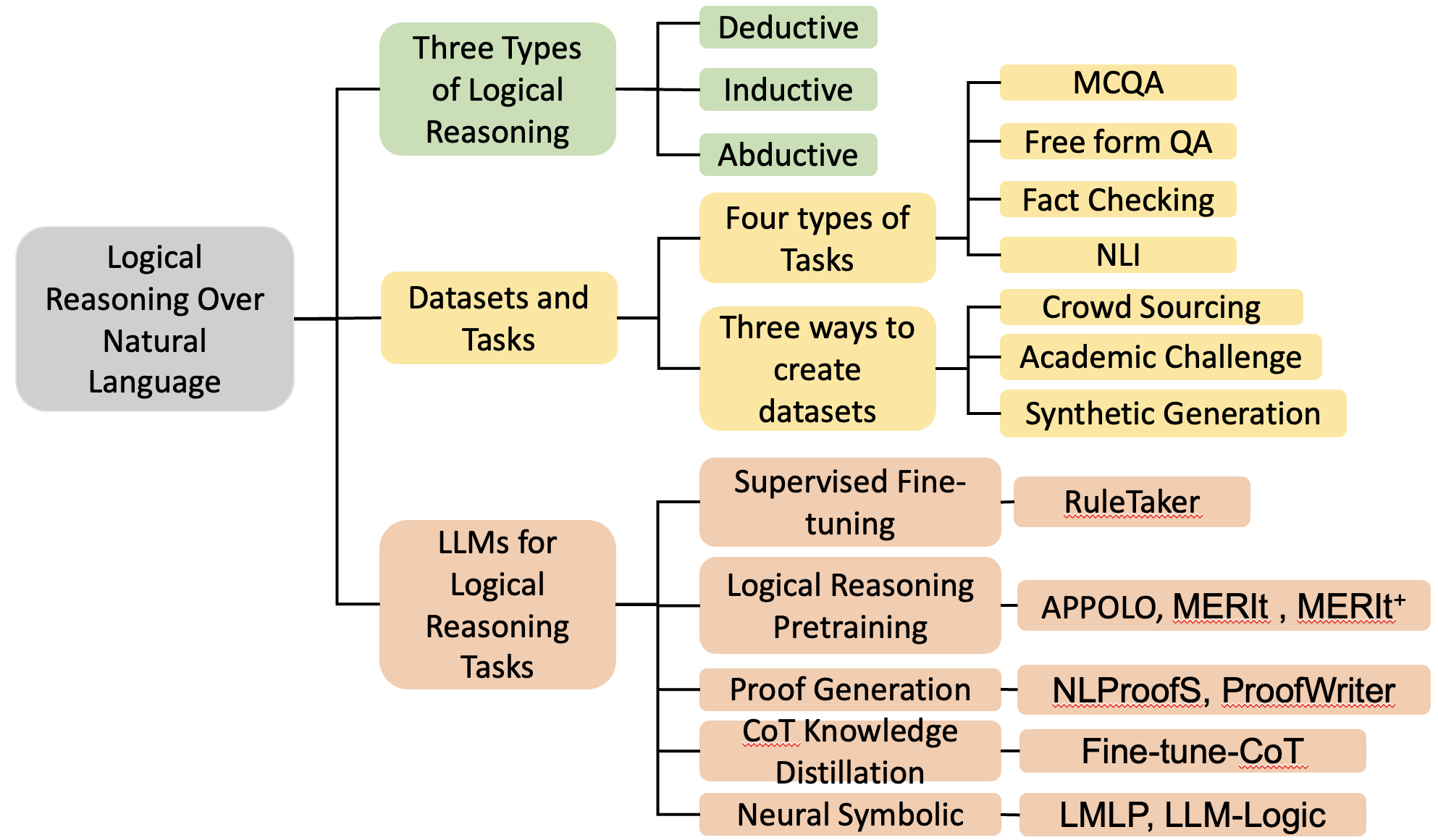}
    \caption{Logical Reasoning Survey: Datasets and Language Model Application.}
    \label{fig:survey}
\end{figure}

Logical reasoning enables humans to \textit{deduce} new conclusions, \textit{learn} (or \textit{induce}) new abstractions and provide plausible explanations (or \textit{abduce}) to various phenomena in the universe. The discovery of black holes serve as a compelling example of the power of logical reasoning. Scientists like Stephen Hawking used the principles of relativity and quantum mechanics to predict their existence and properties. Through formalization and logical deductions alone, they predicted the presence of these mysterious cosmic entities; fifty years before it was confirmed by MIT researchers through physical experiments.  
From early days, Artificial Intelligence (AI) researchers therefore have devoted significant attention towards developing logical reasoning  abilities~\citep{mccarthy1989artificial,colmerauer1996prolog}. Early efforts revolved around the design of formal logical languages to encapsulate rules and facts (domain knowledge), alongwith automated inference engines~\citep{lifschitz2019answer}. This paradigm necessitated expert understanding of the target domain, and, understanding of syntax and semantics of a logical language; which made knowledge \textit{representation} and \textit{acquisition} hard, and an expert-driven endeavor. These challenges motivated the contemporary researchers to progressively turn towards addressing logical reasoning tasks~\citep{Clark2020ruletaker,tian2021logicnli,han2022folio} by employing transformer-based~\citep{Vaswani2017transformer} pre-trained language models~\citep{Devlin2019BERTPO,brown2020language}.   


The Language models (LMs) that are pretrained using objectives such as the mask language modeling~\citep{Devlin2019BERTPO} and next word prediction~\citep{brown2020language} enables them to acquire adequate syntax and semantics of language, alongside commonsense knowledge. These language models excel in numerous natural language understanding tasks, owing to the unsupervised pretraining on a vast array of unstructured text data.
However, it is unclear if the current pretraining objectives are sufficient for the models to infer logical reasoning because this involves understanding structure; coupled with inductive, deductive, and abductive reasoning skills. This question has drawn intense attention and inspired different research directions to examine if LMs can learn logical reasoning ability \citep{DBLP:journals/corr/abs-2307-02477,DBLP:journals/corr/abs-2307-13702,Clark2020ruletaker,joshi2020taxinli}. 
For instance, \citet{Clark2020ruletaker} shows that pre-trained language models can serve as a ``soft-reasoner" based on their near-perfect performance on synthetic datasets. \citet{creswell2022selection} shows that large LMs are few-shot logical reasoners. 
On the other hand, \citet{liu2020logiQA,joshi2020taxinli,han2022folio} show that logical reasoning remains challenging for language models. Furthermore, \citet{DBLP:journals/corr/abs-2307-02477,DBLP:journals/corr/abs-2307-13702} shows that LLMs maybe retrieving or reciting previously seen facts and steps, instead of actually reasoning. 
\citet{liu2023evaluating} shows that while ChatGPT and GPT-4 generally perform well on some benchmarks, their performance noticeably diminishes on new or out-of-distribution datasets.

To understand, and benchmark the progress of logical reasoning abilities in the era of language models, we offer four contributions. 1) First, we provide a concise \textbf{survey} of the role that reasoning plays within current language models. 2) Based on the insights from our survey, we assemble a logical reasoning \textbf{benchmark} termed as \textsc{LogiGLUE} \textcolor{black}{with tasks where conclusions can be drawn through deductive, inductive, or abductive reasoning based solely on the information provided.} \textsc{LogiGLUE} demonstrates three distinguishing characteristics. \underline{First}, it encompasses diverse logical reasoning tasks evaluation, ensuring a comprehensive assessment of a model's performance across varied logical paradigms. \underline{Second}, the unique format of each dataset within \textsc{LogiGLUE} simplifies both training and evaluation processes, facilitating swift integration into research workflows. \underline{Thirdly}, researchers can easily compare with established baselines, and the \textsc{LogiGLUE} offers the flexibility to seamlessly integrate new datasets in the future, ensuring its lasting relevance in logical reasoning evaluation.
3) 
Drawing on multi-task learning and instruction-fine-tuned models, we trained Flan-T5~\citep{chowdhery2022palm} seq2seq models using \textsc{LogiGLUE}'s data and evaluated the efficacy of various fine-tuning strategies. Our findings reveal that approaches involving multi-task fine-tuning and chain-of-thought knowledge distillation lead to superior performance.
4) We comprehensively assess multiple LLMs performance on LogiGLUE, and find that LLMs perform the best at abductive reasoning, followed by deductive reasoning, while they are least effective at inductive reasoning. 




\section{A Concise Survey of Logical Reasoning in NLP}

\begin{figure*} 
    \centering
    \includegraphics[width=\linewidth]{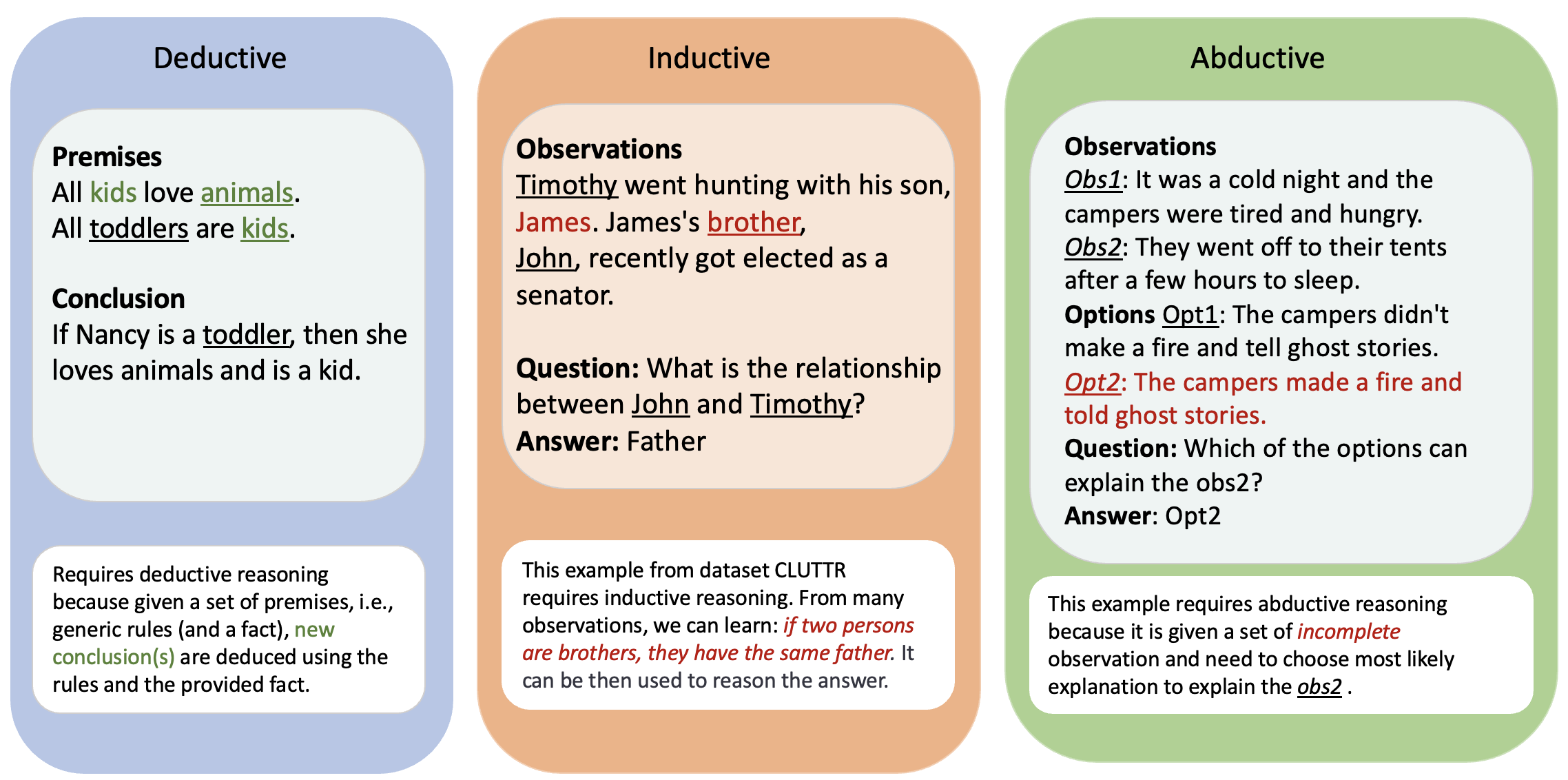}
    \caption{Examples (top) of three types of logical reasoning and explanations (bottom) correlating each example with its respective reasoning type.}
    \label{fig:reasoning_examples}
\end{figure*}

\subsection{Three Types of Logical Reasoning}
\label{sec:reasoning_types}
\textbf{Deductive Reasoning.} In this predominant form of reasoning, we start with a set of premises which can be facts or rules, and derive a specific conclusion based on the available premises with a valid logical derivation path. In short, deductive reasoning derives specific conclusion(s) from generic observation(s)~\citep{byrne2019human}.
There are two characteristics of a deductive reasoning system, \textit{validity} and \textit{soundness}. A conclusion is \textit{valid} if and only if it is fully supported by the premises irrespective of the factuality of the premises. A conclusion is \textit{sound} if and only if it is valid and the premises are true. For example in Figure~\ref{fig:reasoning_examples}, the conclusion ``All kids love animals'' is valid but is not sound.  
Most of the synthetic deductive reasoning datasets such as RuleTaker~\citep{Clark2020ruletaker} has valid conclusions, but may not be sound as the synthetically generated rules in the premises may be untrue in the real world. Datasets such as PrOntoQA~\citep{abulhair} offer a broader view, by sourcing the premise rules from a true, false and a fictional ontology.

\textbf{Inductive Reasoning.}  For inductive reasoning, one starts with a set of observations, and derives a general conclusion that is merely true, but not certain~\citep{heit_2007,sauce2017inductive}.  In contrast to deductive reasoning, inductive reasoning is a bottom-up reasoning process which starts from specific observations and derives a generic conclusion. 
Many Knowledge graph completion task requires inductive reasoning such as WN18RR\footnote{Here, we exclude this task as we are more interested in natural language input. In this paper, we do not discuss about knowledge graph completion tasks since most of them are not in natural language forms.}. To apply inductive reasoning, one usually relies on a large number of observations (both positive and negative in support or against an induced rule). Since large language models are pretrained on large amount of free-text, it learns several generic patterns or conclusions, therefore reasoning inductively (even if the rules may not be represented symbolically or a human-readable fashion)~\citep{han2023inductive}. 
Commonsense reasoning tasks in NLP may require both inductive and deductive reasoning. 





\textbf{Abductive Reasoning.} Abductive reasoning typically begins with an incomplete set of observations and proceeds to derive most likely explanations for the observations to be true~\citep{paul1993approaches,hobbs1993interpretation}.
Similar to inductive reasoning, this also involves uncertainty, as there can be different explanations. While deductive reasoning is a process from known facts or rules to derive a new conclusion, while abductive reasoning is from an observation to a ``guess'' of what can be the reason to cause the observation. 
It is used more often in our daily decision-making, such as medical diagnoses based on an incomplete set of symptoms.

\textbf{Remark.} Both inductive and abductive reasoning inherently encompass uncertainty. In fact, deductive reasoning can also operate within the realm of uncertainty~\citep{problograedt2007,mlnrichardson2006,lee2016weighted,psljmlr2017,lee2018strong}. Such reasoning paradigm uses ``soft rules'' to indicate the likelihood of a rule being true rather than its absolute truth. Consequently, conclusions derived may carry probabilistic true/false values.
\textit{Reasoning under uncertainty} is particularly useful because the real world is inherently unpredictable and full of unknown variables.  
While uncertainty is important, we found that most of the evaluation datasets are under the assumption that rules are unequivocally true, one exception is Rulebert~\citep{saeed2021rulebert} which deviates by attributing weight values to each rule.

\subsection{Logical Reasoning Tasks and Datasets}

There exists various tasks connected to logical reasoning, including counterfactual reasoning~\citep{tandon2019wiqa,frohberg2022crass}, spatial reasoning~\citep{mirzaee2021spartqa}, and mathematical reasoning~\citep{mishra2022numglue}. These tasks often demand additional knowledge, such as understanding spatial relationships and requiring equation calculations, and therefore are not included in the following discussion. 

\subsubsection{Four Types of Tasks}
\label{sec:four_types}
Our survey reveals that logical reasoning datasets predominantly fall into four task formats: Multiple-choice QA (MCQA), Free-Form QA (FFQA), fact-checking (FV), and Natural Language Inference (NLI). MCQA and FFQA involve question-answering with varied answer formats~\citep{yu2019reclor,liu2020logiQA,weston2015babi,banerjee2020can}. Fact-Checking requires validating the truth value of a fact against given information~\citep{Clark2020ruletaker,saeed2021rulebert,he2021winologic}, while NLI entails analyzing a premise and hypothesis to determine their logical relationship~\citep{tian2021logicnli}.

\subsubsection{Dataset Creation Approach}
\label{sec:dataset_creation}
\noindent\textbf{Human Annotation via Crowdsourcing.}
This manner offers benefits like diverse linguistic structures and complex task designs. We find that most of the logicial NLI datasets are created in this manner. 
However, it's costly and often introduces biases, impacting dataset quality and inflating neural model accuracy scores~\citep{yu2019reclor}. 
Additionally, creating logical reasoning datasets requires expertise, adding to the challenge.

\noindent\textbf{Extraction from Academic Challenges.} While crowdsourcing workers struggle to create complex logical reasoning questions due to the need for extensive training, standardized test questions can serve as a source for these datasets after preprocessing~\citep{yu2019reclor,liu2020logiQA}. However, the scope of these exams is narrow and results in smaller dataset sizes.

\noindent\textbf{Synthetic Generation.} Such a method offers a more efficient way to create large datasets compared to manual methods.
It includes simulation-based approaches~\citep{weston2015babi} and rule-based methods~\citep{Clark2020ruletaker,saeed2021rulebert,banerjee2020can}, where logic programs are transformed into natural language. However, this method may produce rules or facts with limited real-world applicability and simplistic language.

\subsection{Language Models for Logical Reasoning over Natural Language}
Recent studies on language models (LMs) for logical reasoning tasks have shown promising developments. Large language models (LLMs) demonstrate human-like abstract reasoning skills~\citep{dasgupta2022language}. \citet{creswell2022selection} introduced a selection-inference pipeline for contextual fact selection and question decomposition. \citet{wei2022chain} highlighted the capability of language models in chain-of-thoughts (CoT) reasoning, aiding in tasks like mathematical reasoning. The next section summarizes five key trends in using LMs for logical reasoning.

\paragraph{Supervised Finetuning.}
Fine-tuning language models on downstream tasks is a standard approach for logical reasoning tasks~\citep{Clark2020ruletaker,liu2021logiqa,tian2021diagnosing,saeed2021rulebert,han2022folio,chan2023self}. This method typically employs moderate-sized models like BERT, GPT2, RoBERTa, and XLNet~\citep{devlin2019bert,radford2019language,liu2019roberta,Yang2019xlnet}. Transformer-based models outperform others like LSTM, partly due to their inherent commonsense and logical reasoning capabilities~\citep{huang2019cosmos,Clark2020ruletaker}. Larger models tend to perform better, indicating a correlation between model depth and reasoning complexity~\citep{he2021winologic}. Despite high in-domain performance, these models often lack generalization across different reasoning depths, language levels, and domains~\citep{Clark2020ruletaker,richardson2021nlsat,tafjord2021proofwriter,banerjee2020can}, suggesting they may learn patterns rather than underlying reasoning skills~\citep{zhang2022paradox}.

\paragraph{Logical Reasoning  Pretraining.} 
Language models learn syntax, semantics, and world knowledge through tasks like next-word prediction or masked language modeling, but this doesn't ensure logical reasoning skills. To address this, researchers are developing logical reasoning-oriented pretraining tasks. APOLLO \citep{sanyal2022apollo} uses two tasks: selective masked language modeling (s-MLM), masking logical words, and entailment classification for relationships in masked sentences. MERIt~\citep{Fangkai2022} introduces a meta-path-guided task, converting documents into graphs for self-supervised learning, distinguishing positive from negative entity-linked sentences. MERIt$^+$~\citep{jiao2023logicllm} evolves this by optimizing the probability of positive sentences, moving away from contrastive learning.

\paragraph{Proof Generation.}

Proof generation is more challenging than answer generation but yields better out-of-domain or unseen depth reasoning performance~\citep{saha2020prover,tafjord2021proofwriter}. NLProofS, introduced by \citet{kaiyuy}, generates step-by-step proofs using a prover, a verifier for step validity, and an algorithm to retrieve the highest validity score proof. ProofWriter~\citep{tafjord2021proofwriter} employs T5 models for proof generation, either predicting the entire proof sequence at once or iteratively generating conclusions. Proof strategies typically involve forward chaining (starting with facts and deriving inferences) or backward chaining (starting with a target and breaking it into sub-goals). \citet{kazemi2022lambada} found backward chaining more effective for LLMs in deductive logical reasoning tasks.

\paragraph{CoT Knowledge Distillation.}
While previous methods depend on proof annotations in datasets, it's shown that large language models (LLMs) can generate step-by-step reasoning, akin to proofs~\citep{abulhair,liu2023logicot}. \citet{NamgyuHo2022} introduced the Fine-tune-CoT (chain-of-thought) approach, which involves a three-step process. First, a large teacher model generates multi-step explanations for complex queries, which are then filtered for accuracy. Second, these explanations form reasoning samples, combining the question, rationale, and answer for comprehensive prompts. Finally, a student model is fine-tuned with these samples to enhance its reasoning capabilities. However, LLMs sometimes struggle with planning proofs, leading to incomplete proofs and potential inaccuracies.

\paragraph{Neural Symbolic.} 
Pretrained Language Models (PLMs) show promise in natural language reasoning but have inconsistent failures. In contrast, symbolic solvers excel in symbolic reasoning tasks but struggle with diverse natural language inputs. A viable approach is using LLMs to convert natural language inputs into symbolic programs for symbolic solvers, reducing LLM unfaithfulness~\citep{pan2023logic}. \citet{hanlingzhang2022} use LLMs as Logic Programmers (LMLP) to reason over knowledge bases with Prolog's backward chaining, outperforming CoT in deductive reasoning. \citet{pan2023logic} developed Logic-LM for deductive reasoning and constraint programming tasks, translating natural language to formal language for symbolic engines. This paradigm is explored further, with LLMs as planners and external tools for execution~\citep{lu2023chameleon,sumers2023cognitive,paranjape2023art,guan2023leveraging,schick2023toolformer} and LLMs augmented with inputs from Logic provers \cite{OGLZ_LINC_2023}. \citet{feng2023language} addressed LLMs' syntax parsing issues impacting solver accuracy by fine-tuning an LLM with an instruction-tuning dataset, thus improving parsing accuracy.

\subsection{Survey Summary} 



Our survey outlines the current datasets covering three logical reasoning types across four task formats and how dataset curation impacts difficulty levels. We've highlighted five methods for employing LLMs in these reasoning tasks. These insights lay the groundwork for future research, guiding improvements in model performance and dataset creation. Next, we introduce a logical reasoning benchmark, aligned with established benchmarks like SuperGlue~\citep{wang2019superglue}, BigBench~\citep{srivastava2023beyond}, and Unicorn~\citep{lourie2021unicorn}, to thoroughly assess system capabilities. Unlike existing reasoning benchmarks such as CROW~\citep{ismayilzada-etal-2023-crow}, our focus is evaluating diverse logical reasoning ability with limited need for implicit/external knowledge.


\section{\textsc{LogiGLUE}: General Logical Reasoning Benchmark} 

\begin{table*}[t]
\centering
 \resizebox{\linewidth}{!}{
\begin{tabular}{l|c|c|c|c|c|c}
    \toprule
     {\bf Dataset}  &  {\bf Train size } &  {\bf Dev size } & {\bf Test size } & {\bf  Synthetic}& {\bf Task Type}  & {\bf Reasoning Type }\\
     \toprule
    \multicolumn{3}{l}{\bf Fine-tuning Set}\\
    \bottomrule
        $\alpha$ARCT~\citeyear{niven2019probing}   &  2420 & 632 & 888 & \xmark  & MCQA  & Abductive \\
        $\alpha$NLI~\citeyear{bhagavatulaabductive} & 169,654 & - & 1532 & \xmark & NLI & Abductive \\
        AbductionRule-Animal ~\citeyear{ignatiev2019abduction} & 23,100 & 3,300 & 6,600 & \cmark & FFQA & Abductive \\
        LogicNLI~\citeyear{tian2021diagnosing}   & 16,000 & 2,000 & 2000 & \cmark & NLI & Deductive \\
        ProofWriter~\citeyear{tafjord2021proofwriter}   & 69,814 & 10,158  & 20,058 & \cmark & FV & Deductive \\
        Rulebert-Union~\citeyear{saeed2021rulebert}   &  56,000 & 4,666&9,334 & \cmark & FV & Deductive \\
        FOLIO~\citeyear{han2022folio} & 1004& 204 & 227 & \xmark & FV & Deductive \\
        ANLI~\citeyear{nie2020adversarial}   & 162,865 &  3,200 & 3,200 & \xmark & NLI & Deductive  \\
        CLUTTR-Robust~\citeyear{sinha2019clutrr}  &  10,100 & - & 144 & \cmark & FFQA & Inductive \\
        LogiQA~\citeyear{liu2021logiqa}   &  7,376 & 651 & 651& \xmark & MCQA & Mixed  \\

        
    \toprule
    \multicolumn{3}{l}{\bf Leave-out Testing Set}\\
    \bottomrule
        AbductionRule-person ~\citeyear{ignatiev2019abduction}  & - & - & 4,864 & \cmark & FFQA & Abductive \\
        bAbi~\citeyear{weston2015towards} &  - & - & 5000 & \cmark & FFQA & Deductive  \\
        Bird-Electricity~\citeyear{tafjord2021proofwriter} &-  &  - & 5270 & \xmark & FV & Deductive \\
        NatlLang~\citeyear{tafjord2021proofwriter}  &-   & -  & 8,008 & \xmark & FV & Deductive \\
        Winologic~\citeyear{he2021winologic}  &-  &  -  & 562  & \xmark & FV & Deductive \\
        WaNLI~\citeyear{liu2022wanli} &- & - & 5000 & \cmark & NLI & Deductive \\
        Rulebert-Union~\citeyear{saeed2021rulebert}  & - & - &5000 & \cmark & FV & Deductive \\
        PrOntoQA~\citeyear{SaparovHe2023}  & - & - & 200  & \xmark & MCQA & Deductive 
        \\    
        BigBench~\citeyear{srivastava2022beyond} & - & - & 1300  & \xmark & FFQA & Deductive \\ 
        BigBench~\citeyear{srivastava2022beyond}  & - & - & 32  & \xmark & FFQA & Inductive \\
        bAbi~\citeyear{weston2015towards} &  - & - & 5000 & \cmark & FFQA & Inductive   \\
        CLUTTR-Systematic~\citeyear{sinha2019clutrr}   &  - & - & 10100 & \cmark & FFQA & Inductive \\
        ReClor~\citeyear{yureclor}   & -  & - & 500 &  \xmark & MCQA & Mixed  \\
        LogiQA 2.0~\citeyear{liu2023logiQA}  & - & - & 3238  & \xmark & NLI & Mixed 
        \\
    \bottomrule
    \end{tabular}
    }
\caption{Statistics of In-domain (IID) and out-of-domain (OOD) datasets of LogiGLUE benchmark.}
\label{tab:datasets}
\end{table*}

Various studies have reported differing evaluations of language models' logical reasoning skills~\citep{Clark2020ruletaker,joshi2020taxinli,han2022folio}, attributing inconsistencies to the varied evaluation benchmarks and task formats employed. In response, we have compiled \textsc{LogiGLUE}, a comprehensive benchmark to uniformly assess the logical reasoning capabilities of language models. This benchmark consists of 24 datasets, segmented into a fine-tuning set (10 datasets) and a testing set left out of training (14 datasets), detailed license information for which can be found in Appendix~\ref{apd:license}. The fine-tuning set is designed for training on logical reasoning tasks, while the leave-out testing set aims to test the model's ability to generalize to new, unrepresented domains beyond the fine-tuning dataset.

\textbf{Diversity.}  Table~\ref{tab:datasets} demonstrates that \textsc{LogiGLUE} spans the entire spectrum of logical reasoning by encompassing all three primary reasoning types. Furthermore, \textsc{LogiGLUE} incorporates datasets across four task categories and three methodologies of creation. Although we aim for varied reasoning types in both fine-tuning and leave-out testing sets, we observe a trend in the literature towards more datasets designed for deductive reasoning. This may be attributed to the feasibility of template-based generation for deductive reasoning, in contrast to the complexities involved in creating large-scale datasets for inductive and abductive reasoning.

\textbf{Unique Format.} 
Existing practices standardize diverse datasets into a consistent format, such as question answering or NLI styles~\citep{mishra2022cross,lourie2021unicorn}. 
A consistent format cross diverse datasets not only simplifies the training and testing process but also beneficial to build a single model that are capable of performing different tasks. 
Based on this practical reason, we convert all 24 datasets into a Sequence2Sequence format. Specifically, the input of the MCQA/FV/NLI tasks include a context, question, and answer choices, while FF tasks do not have answer options. 
For FV, statements serve as questions with true/false answer choices, and NLI tasks offer natural, contradictory, and entailment options. For all four types of tasks, the outputs are the answer string.

\textbf{Comparison with Other Logic Reasoning Benchmark.} While various initiatives aim to create a benchmark for assessing logical reasoning~\citep{teng2023glore,helwe2022logitorch,anonymous2023logicbench,ismayilzada-etal-2023-crow}, \textsc{LogiGLUE} sets a new standard by offering a uniquely formatted, extensive training set. It encompasses an array of reasoning types and provides a well-organized, varied categorization of datasets. We compare with each benchmark in Table~\ref{tab:other_logi_datasets}. Details discussion is given in Appendix~\ref{apd:other_benchmark}.
\begin{table}[t]
\centering
 \resizebox{0.9\linewidth}{!}{
\begin{tabular}{c|c|c|c|c|c}
    \toprule
    Benchmark & {\# Tasks} &  {Train}  & {In-Domain Test} & {Leave-out Test} & Annotation of Reasoning Type\\
    \toprule 
    GLoRe~\citeyear{teng2023glore} & 12 & \xmark & \cmark & \cmark & \xmark\\
    LogiTorch~\citeyear{helwe2022logitorch} & 14 & \cmark & \cmark & \xmark & \xmark\\
    LogiBench~\citeyear{anonymous2023logicbench}  & 3 & \cmark & \cmark & \xmark & \cmark \\
    CROW~\citeyear{ismayilzada-etal-2023-crow} & 6 & \xmark & \cmark & \cmark & \xmark\\ 
    \textsc{LogiGLUE} (ours) & \textbf{24} & \cmark & \cmark & \cmark & \cmark\\
    \bottomrule
    \end{tabular}
    }
\caption{Compare \textsc{LogiGLUE} with other logical reasoning benchmarks. Train: include training or not; Test: include training or not; OOD: whether include out-of-domain evaluation, Annotation: Annotation of Type of Reasoning, }
\label{tab:other_logi_datasets}
\end{table}

\section{Experiments and Results} \label{sec:experiments}

We mainly design two sets of experiments. 
The first focus on fine-tuning a relative small LM for logical reasoning tasks to evaluate the impact of various fine-tuning techniques. The second experiment tests pretrained language models using the leave-out testing set, aiming to assess and compare their performance across different reasoning types.

\subsection{Fine-tuning Performance}
We selected Flan-T5-large~\citep{chowdhery2022palm} as our base model for training due to two pivotal reasons. Firstly, Flan-T5 stands as an instruction-fine-tuned iteration of T5, exhibiting enhanced performance when compared to its peers. Secondly, Flan-T5's manageable size renders it to be trainable on a machine that is conducive to an academic setting. 

\textbf{Single Task Fine-Tuning.} 
We fine-tune Flan-T5-large on each dataset and report average accuracy in Table~\ref{tab:single_domain_result}. The model exhibits superior performance on synthetic data compared to hand-crafted alternatives as demonstrated by the observation that the top 5 performances are predominantly synthetic.
This trend holds while considering the ANLI dataset, which, despite having a more substantial training set than its synthetic counterparts, yielded inferior results. 
This suggests that there might be some heuristic patterns exhibited in the synthetic datasets that the model can easily learn rather than learning the underlying logic.
\begin{table}[t]
\centering
 \resizebox{0.8\linewidth}{!}{
\begin{tabular}{l|c|c|c }
    \toprule
     {\bf Model}  & {\bf Dataset Creation Approach} & \begin{tabular}[c]{@{}l@{}}\textbf{Test Split of}\\ \textbf{Corresp. Dataset}\end{tabular} &  \begin{tabular}[c]{@{}l@{}}\textbf{LogiGLUE Avg.}\\ \textbf{In-Domain Perf.}\end{tabular} \\
     \toprule
     \multicolumn{4}{c}{\textbf{Single-Task Model}}\\
     \midrule
     Flan-T5 & \multirow{6}{*}{Synthetic} & \\
       $\;+$ AbductionRule-Animal & & {100.00} & 50.88\\ 
        $\;+$ Rulebert-Union &  & 99.69 & 34.61\\
        $\;+$ ProofWriter & & 99.42 & 23.42 \\
        $\;+$ CLUTTR-Robust & & {97.22} & 50.17 \\   
        $\;+$ LogicNLI  & & 82.60 & 31.49\\
        \midrule
        & \multirow{5}{*}{\begin{tabular}[c]{@{}l@{}}{~~~~Crowd Sourcing}\\ {Academic Challenges}\end{tabular}} & \\
        $\;+$ $\alpha$NLI & & 78.26 & 44.00 \\
        $\;+$ $\alpha$ARCT &  & 72.31 & 45.02\\
        $\;+$ FOLIO & & 66.66 & 46.36 \\
        $\;+$ ANLI & & {61.16} & 21.03\\  
        $\;+$ LogiQA & & 37.94 & 43.32 \\
        \midrule 
        \multicolumn{4}{c}{\textbf{Multi-Task Model}}\\
     \midrule
        LogiT5 & - & - & \textbf{80.88}\\
    \bottomrule
    \end{tabular}
    }
\caption{Performance of fine-tuned Flan-T5 (large) on each single dataset, followed by average performance over LogiGLUE IID Testing datasets.}
\label{tab:single_domain_result}
\end{table}

\begin{table}[t]
\centering
 \resizebox{0.5\linewidth}{!}{
\begin{tabular}{l|c|c|c}
    \toprule
     {\bf Dataset}  &{\bf Training Size}  &  {\bf Single-Task Model} &  {\bf LogiT5} \\
     \toprule
        $\alpha$ARCT & 2420 & 72.31 & \textbf{77.22}  \\
        FOLIO & 1004 & 66.66 & \textbf{74.02}\\
        \midrule
        $\alpha$NLI  & 169,654 & \textbf{78.26}  & 76.37 \\
        ANLI  & 162,865 & \textbf{61.16} & 59.53\\  
    \bottomrule
    \end{tabular}
    }
\caption{Performance of Single Task and LogiT5 model: the first block is low resource domain and the second block is high resource.}
\label{tab:low_high_resouce}
\end{table}

\textbf{Multitask Fine-tuning.} 
We fine-tune the Flan-T5 Large on all in-domain datasets utilizing a weighted sampling technique to accommodate for the unbalanced size of the training datasets.
While the complexity of both multi-task and single-task models is comparable, the multi-task model demonstrates a significantly higher average performance. As indicated in Table~\ref{tab:single_domain_result}, the LogiT5 model surpasses each single-task model by a substantial margin, ranging between 30-60\%. 
Multi-task training presents a distinct advantage over single-task models, particularly in scenarios involving low-resource data as shown in Table~\ref{tab:low_high_resouce}.
This is supported by research~\citep{parmar2022boxbart,luo2022choose}. Specifically, in cases of limited training data (i.e., limited to 1/2 K training samples, first block results in Table~\ref{tab:low_high_resouce}), LogiT5 tends to outperform single-task models, likely due to its ability to transfer skills or knowledge acquired from other datasets to enhance performance in low-resource domains. This is evident in LogiT5's enhanced performance, showing improvements of 5\% and 8\% in the $\alpha$ARCT and FOLIO tasks, respectively.
In contrast, for tasks with ample training data, such as the $\alpha$NLI and ANLI datasets (second block results in Table~\ref{tab:low_high_resouce}), multi-task training does not offer additional benefits. This could be attributed to the fact that a large training set already provides a robust learning environment for the model, making the added value of multi-task training negligible. This trend highlights a significant constraint of multi-task training in situations where individual datasets are already extensive, limiting its effectiveness.



\paragraph{Fine-tuned LogiT5 on Single Dataset.}
Here, we further fine-tune LogiT5 on each dataset. However, upon analyzing the performance, we did not observe notable advantages from this additional fine-tuning even though small margin gains are achieved, which only yield 0.4\% average improvement. 
The results of each dataset is given in Table \ref{tab:indomain_result} in Appendix. 
This suggests that LogiT5 has likely already learned the majority of knowledge from these tasks.

\paragraph{CoT Distillation Fine-tuning.} 
As shown by previous work~\citep{NamgyuHo2022}, distill the chain-of-thoughts from a large model to a small student model boosting the performance of the student model. We apply such CoT finetuning strategy and conduct experiments on LogiQA, identified as the most challenging task, by distilling the CoT from LLama-chat-7B to Flan-T5 Large. Initially, we generated a single answer for each question, retaining only the samples where the predicted answer was correct, resulting in approximately 3K valid samples (\textbf{Type 1}).
Alternatively, we created 10 answers for each question and preserved the samples with at least one correct predicted answer, which generated a unique set of 6K questions (\textbf{Type 2}). It is worth noting that some questions offered multiple correct reasoning paths. In such cases, we either opted for a singular path or utilized all available paths, the latter approach amassing a total of 15K training samples (\textbf{Type 3}). We trained the Flan-T5 model utilizing datasets consisting of 3K, 6K, and 15K samples, derived from the generated CoT, with the results delineated in Table \ref{tab:logiQA_cot}. Our findings indicate that the training with 3K and 6K samples did not enhance the CoT's fine-tuning efficacy. However, an increased dataset size of 15K samples facilitated a 4\% improvement in performance, suggesting that CoT distillation becomes more beneficial with a larger volume of data.
\begin{table}[t!]
\centering
 \resizebox{0.5\linewidth}{!}{
\begin{tabular}{c|c|c|c|c}
    \toprule
    Data & {Type 0} &  {Type 1 }  & { Type 2 } & {  Type 3 } \\
     \toprule
    Acc & 0.37 & 0.38 & 0.37 & \textbf{0.41} \\
    \bottomrule
    \end{tabular}
    }
\caption{Performance of Flan-T5 trained with CoT fine-tuning on the LogiQA dataset. Type 0 is without using CoT 7K data, Type 1 is with CoT 3K data, Type 2: CoT 6K data, Type 3: CoT 15K.}
\label{tab:logiQA_cot}
\end{table}

\subsection{Evaluations of Pretrained LM and LogiT5 on Leave-out Testing Set}
\begin{table}[!htb]
\centering
 \resizebox{0.95\linewidth}{!}{
\begin{tabular}{l|c|c|c|c|c|c}
    \toprule
     \textbf{Type of Reasoning} & {\bf Dataset}  & \textbf{ GPT4} & \textbf{GPT3.5} & \textbf{LLaMA-Chat-7B}	& \textbf{Flan-T5}	& \textbf{LogiT5} \\
     \toprule
    \multirow{8}{*}{~~~~~~Deductive} & bAbi-deductive	& \textbf{97.39} & {87.80} & 	69.10  & 	38.90 & 	65.00\\ 
    & Bird-Electricity & 64.61 & 	49.24 &	50.94 &	42.12&	\textbf{66.32} \\
    
    & NatLang	& 52.46 & 50.94	& 48.80 & 	57.00& 	\textbf{70.24}\\
    & Winologic & \textbf{88.39} & 45.53 &	41.96 & {66.07} & 60.71 \\
    & WaNLI & \textbf{65.20} & 45.02 &	31.50  & 49.90 & {61.90}\\	
    & Bigbech-deduction & \textbf{84.38} & 40.00 & 43.00 &	{46.92} &	33.84 \\
    & Rulebert-Union	& 62.40 & 47.20 & 	\textbf{65.00}  & 24.80& {61.90} \\ 
    & prOntoQA & \textbf{86.00} & 28.00 & 	{60.20}  & 	6.50 & 	33.50 \\
    & \textit{Avg.}	& \textbf{73.59} & 49.21 & 48.05 & 41.52 & 56.67 \\ 
    \midrule
    \multirow{4}{*}{~~~~~~Inductive} & CLUTTR-Systematic & 9.00 & 8.11 & 	8.60& 15.75  & 	\textbf{100	}\\
    & Bigbech-logical-args & \textbf{72.30} &	40.63 &	37.50&	{50.00 }  & 	40.63 \\
    & bAbi-inductive	& 1.30 & 5.40 & 30.80  & \textbf{56.49} & 	12.90 \\ 
    & \textit{Avg. } & 31.55 & 18.04 & 25.63 & 40.74 & \textbf{51.17} \\
    \midrule
    ~~~~~~Abductive & AbductionRule & \textbf{98.76} & 	68.20 & 	52.47	& 3.76 & {96.10}\\
    \midrule
    \multirow{3}{*}{~~~~~~Mixed}& ReClor 	& \textbf{76.00} & {61.00}	&41.00 &	42.00 & 	47.00 \\
    & LogiQA 2.0	& \textbf{89.95} & {63.83} & 	61.36 & 	49.30  & 	45.20\\
    & \textit{Avg.} & \textbf{82.98} & 62.42 & 51.18 & 45.65 & 46.10\\    
    \bottomrule
    \end{tabular}
    }
\caption{Performance of Different Language Models on Leave-out Testing Set.}
\label{tab:outdomain_result}
\end{table}
We examine three large language models: GPT-4, GPT-3.5-turbo, and LLama-2-chat (7B). We also examine Flan-T5 large and LogiT5, which are relative smaller language models (less than 1B parameters). 
We randomly sample 20\% of the testing sets for evaluation due to the the inference cost of GPT-4 and GPT-3.5-turbo. We provide the results on the full set using the other models in Table~\ref{tab:outdomain_result} in Appendix~\ref{apd:ood}. Our evaluation is based on zero-shot capabilities using chain-of-thought prompting. The prompt is given in the Appendix~\ref{apd:zero-shot-prompt}. 

\textbf{Comparison Across Different Models.} 
Table~\ref{tab:outdomain_result} clearly shows that GPT-4 outperforms other models in overall performance. Remarkably, LLaMA-2-Chat, despite having fewer parameters than GPT-3.5, manages to achieve comparable results. 
In addition, we dissect why GPT-4 and GPT-3 performs worse on some datasets. In our error analysis, we observed that for some tasks, such as CLUTTR and baBi (inductive reasoning), GPT-3 frequently says that no relationship or answer could not be determined. 
Sometimes, there is a flaw in the model's generated reasoning chain, leading to wrong answers. Take bAbi as an example, the given input is: 
Context: Wolves are afraid of mice. Winona is a cat. Mice are afraid of sheep. Sheep are afraid of wolves. Jessica is a sheep. Cats are afraid of wolves. Emily is a mouse. Gertrude is a wolf. Question:  What is Gertrude afraid of?  The ground truth is mouse, however the 
GPT-3 response is ``Based on the context, Gertrude is a wolf. From statement 4, we know that sheep are afraid of wolves. Therefore, Gertrude is afraid of sheep''. To answer the question, we should know what the wolf is afraid of, rather than what is afraid of wolves. This highlights that the model makes an error in understanding the direction of the relationship. 

\textbf{Comparison Across Different Type of Reasoning.}
Table~\ref{tab:outdomain_result} reveals a consistent pattern in the performance of all three LLMs across reasoning types, ranking from strongest to weakest as abductive, deductive, and inductive reasoning. As discussed earlier, abductive reasoning involves inferring the most plausible cause behind an observation, a task at which these LLMs excel due to their extensive training on vast text corpora. Conversely, inductive reasoning requires the extrapolation of patterns from broad observations, a more challenging task for LLMs approached in a zero-shot manner due to insufficient exposure to varied instances required for pattern recognition. Incorporating few-shot in-context learning could potentially enhance inductive reasoning capabilities.

\textbf{Compare Fine-tuned Model with Pretrained Models.}
When we assess the performances of LogiT5 and Flan-T5 as shown in Table~\ref{tab:outdomain_result}, it is evident that LogiT5 generally outperforms Flan-T5, indicating that fine-tuning offers significant benefits to smaller LMs. In comparison to other pretrained large language models (LLMs) like GPT-3 and LLaMA, LogiT5 is comparable or even surpasses them in some datasets, particularly in tasks requiring inductive and abductive reasoning. However, in datasets involving mixed reasoning, LLMs significantly outperform LogiT5. This may arise because datasets like ReClor and LogiQA, both of which are from academic challenges and  publicly accessible, therefore, might have been included in the LLMs' training corpus, allowing these models to recall relevant information.




\section{Conclusion} 
In this study, we concentrate our efforts on a crucial area of research: logical reasoning over natural language. Initially, we offer a survey to provide a thorough comprehension of this domain, emphasizing the role of large language models in addressing this demanding task. Following this, we assemble a benchmark for logical reasoning named \textsc{LogiGLUE}, set to be publicly available to aid forthcoming research. Finally, utilizing \textsc{LogiGLUE}, we fine-tune a language model and asses LLMs logical reasoning capability .

\section*{Limitation} 
Our survey, while comprehensive, focuses on the primary trends in enhancing the logical reasoning capabilities of Large Language Models (LLMs), given the rapid advancements in this field. While it may not encompass every paper on logical reasoning evaluation or methodology, it provides a robust overview of significant developments.
Regarding the \textsc{LogiGLUE} benchmark, it captures the most widely studied datasets at the time of publication. While recent datasets like LogiBench is not included initially, we plan to continuously integrate new logical reasoning datasets into our open-source benchmark for both training and testing, ensuring its relevance and utility in the evolving landscape of LLM research.

\bibliography{custom}

\begin{thebibliography}{91}
\providecommand{\natexlab}[1]{#1}
\providecommand{\url}[1]{\texttt{#1}}
\expandafter\ifx\csname urlstyle\endcsname\relax
  \providecommand{\doi}[1]{doi: #1}\else
  \providecommand{\doi}{doi: \begingroup \urlstyle{rm}\Url}\fi

\bibitem[Anonymous(2023)]{anonymous2023logicbench}
Anonymous.
\newblock Logicbench: Towards systematic evaluation of logical reasoning ability of large language models.
\newblock In \emph{Submitted to The Twelfth International Conference on Learning Representations}, 2023.
\newblock URL \url{https://openreview.net/forum?id=71kocBuhNO}.
\newblock under review.

\bibitem[Bach et~al.(2017)Bach, Broecheler, Huang, and Getoor]{psljmlr2017}
Stephen~H. Bach, Matthias Broecheler, Bert Huang, and Lise Getoor.
\newblock Hinge-loss markov random fields and probabilistic soft logic.
\newblock \emph{J. Mach. Learn. Res.}, 18\penalty0 (1):\penalty0 3846–3912, jan 2017.
\newblock ISSN 1532-4435.

\bibitem[Banerjee et~al.(2020)Banerjee, Baral, Luo, Mitra, Pal, Son, and Varshney]{banerjee2020can}
Pratyay Banerjee, Chitta Baral, Man Luo, Arindam Mitra, Kuntal Pal, Tran~C Son, and Neeraj Varshney.
\newblock Can transformers reason about effects of actions?
\newblock \emph{arXiv:2012.09938}, 2020.

\bibitem[Bhagavatula et~al.(2019)Bhagavatula, Le~Bras, Malaviya, Sakaguchi, Holtzman, Rashkin, Downey, Yih, and Choi]{bhagavatulaabductive}
Chandra Bhagavatula, Ronan Le~Bras, Chaitanya Malaviya, Keisuke Sakaguchi, Ari Holtzman, Hannah Rashkin, Doug Downey, Wen-tau Yih, and Yejin Choi.
\newblock Abductive commonsense reasoning.
\newblock In \emph{International Conference on Learning Representations}, 2019.

\bibitem[Brown et~al.(2020)Brown, Mann, Ryder, Subbiah, Kaplan, Dhariwal, Neelakantan, Shyam, Sastry, Askell, et~al.]{brown2020language}
Tom Brown, Benjamin Mann, Nick Ryder, Melanie Subbiah, Jared~D Kaplan, Prafulla Dhariwal, Arvind Neelakantan, Pranav Shyam, Girish Sastry, Amanda Askell, et~al.
\newblock Language models are few-shot learners.
\newblock \emph{Advances in neural information processing systems}, 33:\penalty0 1877--1901, 2020.

\bibitem[Byrne et~al.(2019)Byrne, Evans, and Newstead]{byrne2019human}
Ruth~MJ Byrne, Jonathan St~BT Evans, and Stephen~E Newstead.
\newblock \emph{Human reasoning: The psychology of deduction}.
\newblock Psychology Press, 2019.

\bibitem[Chan et~al.(2023)Chan, Liu, Chan, Cheng, Song, Wong, and See]{chan2023self}
Chunkit Chan, Xin Liu, Tsz~Ho Chan, Jiayang Cheng, Yangqiu Song, Ginny Wong, and Simon See.
\newblock Self-consistent narrative prompts on abductive natural language inference.
\newblock \emph{arXiv preprint arXiv:2309.08303}, 2023.

\bibitem[Chowdhery et~al.(2022)Chowdhery, Narang, Devlin, Bosma, Mishra, Roberts, Barham, Chung, Sutton, Gehrmann, et~al.]{chowdhery2022palm}
Aakanksha Chowdhery, Sharan Narang, Jacob Devlin, Maarten Bosma, Gaurav Mishra, Adam Roberts, Paul Barham, Hyung~Won Chung, Charles Sutton, Sebastian Gehrmann, et~al.
\newblock Palm: Scaling language modeling with pathways.
\newblock \emph{arXiv preprint arXiv:2204.02311}, 2022.

\bibitem[Clark et~al.(2020)Clark, Tafjord, and Richardson]{Clark2020ruletaker}
Peter Clark, Oyvind Tafjord, and Kyle Richardson.
\newblock Transformers as soft reasoners over language.
\newblock In Christian Bessiere (ed.), \emph{IJCAI}, 7 2020.
\newblock \doi{10.24963/ijcai.2020/537}.
\newblock URL \url{https://doi.org/10.24963/ijcai.2020/537}.

\bibitem[Colmerauer \& Roussel(1996)Colmerauer and Roussel]{colmerauer1996prolog}
Alain Colmerauer and Philippe Roussel.
\newblock The birth of prolog.
\newblock In \emph{History of programming languages---II}. 1996.

\bibitem[Creswell et~al.(2022)Creswell, Shanahan, and Higgins]{creswell2022selection}
Antonia Creswell, Murray Shanahan, and Irina Higgins.
\newblock Selection-inference: Exploiting large language models for interpretable logical reasoning.
\newblock In \emph{The Eleventh International Conference on Learning Representations}, 2022.

\bibitem[Dasgupta et~al.(2022)Dasgupta, Lampinen, Chan, Creswell, Kumaran, McClelland, and Hill]{dasgupta2022language}
Ishita Dasgupta, Andrew~K Lampinen, Stephanie~CY Chan, Antonia Creswell, Dharshan Kumaran, James~L McClelland, and Felix Hill.
\newblock Language models show human-like content effects on reasoning.
\newblock \emph{arXiv preprint arXiv:2207.07051}, 2022.

\bibitem[De~Raedt et~al.(2007)De~Raedt, Kimmig, and Toivonen]{problograedt2007}
Luc De~Raedt, Angelika Kimmig, and Hannu Toivonen.
\newblock Problog: A probabilistic prolog and its application in link discovery.
\newblock In \emph{Proceedings of the 20th International Joint Conference on Artifical Intelligence}, IJCAI'07, pp.\  2468–2473, San Francisco, CA, USA, 2007. Morgan Kaufmann Publishers Inc.

\bibitem[Devlin et~al.(2019{\natexlab{a}})Devlin, Chang, Lee, and Toutanova]{Devlin2019BERTPO}
J.~Devlin, Ming-Wei Chang, Kenton Lee, and Kristina Toutanova.
\newblock Bert: Pre-training of deep bidirectional transformers for language understanding.
\newblock In \emph{NAACL-HLT}, 2019{\natexlab{a}}.

\bibitem[Devlin et~al.(2019{\natexlab{b}})Devlin, Chang, Lee, and Toutanova]{devlin2019bert}
Jacob Devlin, Ming-Wei Chang, Kenton Lee, and Kristina Toutanova.
\newblock {BERT}: Pre-training of deep bidirectional transformers for language understanding.
\newblock In \emph{NAACL}, Minneapolis, Minnesota, June 2019{\natexlab{b}}. ACL.
\newblock \doi{10.18653/v1/N19-1423}.
\newblock URL \url{https://aclanthology.org/N19-1423}.

\bibitem[Fangkai~Jiao(2022)]{Fangkai2022}
Xuemeng Song Liqiang~Nie Fangkai~Jiao, Yangyang~Guo.
\newblock Merit: Meta-path guided contrastive learning for logical reasoning.
\newblock \emph{arXiv preprint arXiv:2203.00357}, 2022.

\bibitem[Feng et~al.(2023)Feng, Xu, Hao, Sharma, Shen, Zhao, and Chen]{feng2023language}
Jiazhan Feng, Ruochen Xu, Junheng Hao, Hiteshi Sharma, Yelong Shen, Dongyan Zhao, and Weizhu Chen.
\newblock Language models can be logical solvers.
\newblock \emph{arXiv preprint arXiv:2311.06158}, 2023.

\bibitem[Frohberg \& Binder(2022)Frohberg and Binder]{frohberg2022crass}
J{\"o}rg Frohberg and Frank Binder.
\newblock Crass: A novel data set and benchmark to test counterfactual reasoning of large language models.
\newblock In \emph{Proceedings of the Thirteenth Language Resources and Evaluation Conference}, pp.\  2126--2140, 2022.

\bibitem[Guan et~al.(2023)Guan, Valmeekam, Sreedharan, and Kambhampati]{guan2023leveraging}
Lin Guan, Karthik Valmeekam, Sarath Sreedharan, and Subbarao Kambhampati.
\newblock Leveraging pre-trained large language models to construct and utilize world models for model-based task planning.
\newblock \emph{arXiv preprint arXiv:2305.14909}, 2023.

\bibitem[Han et~al.(2022)Han, Schoelkopf, Zhao, Qi, Riddell, Benson, Sun, Zubova, Qiao, Burtell, et~al.]{han2022folio}
Simeng Han, Hailey Schoelkopf, Yilun Zhao, Zhenting Qi, Martin Riddell, Luke Benson, Lucy Sun, Ekaterina Zubova, Yujie Qiao, Matthew Burtell, et~al.
\newblock Folio: Natural language reasoning with first-order logic.
\newblock \emph{arXiv preprint arXiv:2209.00840}, 2022.

\bibitem[Han et~al.(2023)Han, Ransom, Perfors, and Kemp]{han2023inductive}
Simon~Jerome Han, Keith~J Ransom, Andrew Perfors, and Charles Kemp.
\newblock Inductive reasoning in humans and large language models.
\newblock \emph{Cognitive Systems Research}, pp.\  101155, 2023.

\bibitem[Hanlin~Zhang1(2022)]{hanlingzhang2022}
Li~Erran Li3 Eric~Xing Hanlin~Zhang1, Yi-Fan~Zhang2.
\newblock The impact of symbolic representations on in-context learning for few-shot reasoning.
\newblock \emph{arXiv preprint arXiv:2212.08686}, 2022.

\bibitem[He et~al.(2021)He, Huang, Liu, and Zhu]{he2021winologic}
Weinan He, Canming Huang, Yongmei Liu, and Xiaodan Zhu.
\newblock {W}ino{L}ogic: {A} zero-shot logic-based diagnostic dataset for {W}inograd {S}chema {C}hallenge.
\newblock In \emph{EMNLP}, November 2021.
\newblock URL \url{https://aclanthology.org/2021.emnlp-main.307}.

\bibitem[Heit(2007)]{heit_2007}
Evan Heit.
\newblock \emph{What Is Induction and Why Study It?}, pp.\  1–24.
\newblock Cambridge University Press, 2007.
\newblock \doi{10.1017/CBO9780511619304.002}.

\bibitem[Helwe et~al.(2022)Helwe, Clavel, and Suchanek]{helwe2022logitorch}
Chadi Helwe, Chlo\'e Clavel, and Fabian Suchanek.
\newblock Logitorch: A pytorch-based library for logical reasoning on natural language.
\newblock In \emph{Proceedings of the 2022 Conference on Empirical Methods in Natural Language Processing: System Demonstrations}, 2022.

\bibitem[Hobbs et~al.(1993)Hobbs, Stickel, Appelt, and Martin]{hobbs1993interpretation}
Jerry~R Hobbs, Mark~E Stickel, Douglas~E Appelt, and Paul Martin.
\newblock Interpretation as abduction.
\newblock \emph{Artificial intelligence}, 63\penalty0 (1-2):\penalty0 69--142, 1993.

\bibitem[Huang et~al.(2019)Huang, Le~Bras, Bhagavatula, and Choi]{huang2019cosmos}
Lifu Huang, Ronan Le~Bras, Chandra Bhagavatula, and Yejin Choi.
\newblock Cosmos {QA}: Machine reading comprehension with contextual commonsense reasoning.
\newblock In \emph{EMNLP-IJCNLP}, November 2019.
\newblock \doi{10.18653/v1/D19-1243}.
\newblock URL \url{https://aclanthology.org/D19-1243}.

\bibitem[Ignatiev et~al.(2019)Ignatiev, Narodytska, and Marques-Silva]{ignatiev2019abduction}
Alexey Ignatiev, Nina Narodytska, and Joao Marques-Silva.
\newblock Abduction-based explanations for machine learning models.
\newblock In \emph{Proceedings of the AAAI Conference on Artificial Intelligence}, volume~33, pp.\  1511--1519, 2019.

\bibitem[Ismayilzada et~al.(2023)Ismayilzada, Paul, Montariol, Geva, and Bosselut]{ismayilzada-etal-2023-crow}
Mete Ismayilzada, Debjit Paul, Syrielle Montariol, Mor Geva, and Antoine Bosselut.
\newblock {CR}o{W}: Benchmarking commonsense reasoning in real-world tasks.
\newblock In Houda Bouamor, Juan Pino, and Kalika Bali (eds.), \emph{Proceedings of the 2023 Conference on Empirical Methods in Natural Language Processing}, pp.\  9785--9821, Singapore, December 2023. Association for Computational Linguistics.
\newblock URL \url{https://aclanthology.org/2023.emnlp-main.607}.

\bibitem[Jiao et~al.(2023)Jiao, Teng, Joty, Ding, Sun, Liu, and Chen]{jiao2023logicllm}
Fangkai Jiao, Zhiyang Teng, Shafiq Joty, Bosheng Ding, Aixin Sun, Zhengyuan Liu, and Nancy~F Chen.
\newblock Logicllm: Exploring self-supervised logic-enhanced training for large language models.
\newblock \emph{arXiv preprint arXiv:2305.13718}, 2023.

\bibitem[Joshi et~al.(2020)Joshi, Aditya, Sathe, and Choudhury]{joshi2020taxinli}
Pratik Joshi, Somak Aditya, Aalok Sathe, and Monojit Choudhury.
\newblock {T}axi{NLI}: Taking a ride up the {NLU} hill.
\newblock In \emph{CoNLL}, November 2020.
\newblock \doi{10.18653/v1/2020.conll-1.4}.
\newblock URL \url{https://aclanthology.org/2020.conll-1.4}.

\bibitem[Kaiyu~Yang \& Chen(2022)Kaiyu~Yang and Chen]{kaiyuy}
Jia~Deng Kaiyu~Yang and Danqi Chen.
\newblock Generating natural language proofs with verifier-guided search.
\newblock \emph{arXiv preprint arXiv:2205.12443}, 2022.

\bibitem[Kazemi et~al.(2022)Kazemi, Kim, Bhatia, Xu, and Ramachandran]{kazemi2022lambada}
Seyed~Mehran Kazemi, Najoung Kim, Deepti Bhatia, Xin Xu, and Deepak Ramachandran.
\newblock Lambada: Backward chaining for automated reasoning in natural language.
\newblock \emph{arXiv preprint arXiv:2212.13894}, 2022.

\bibitem[Lanham et~al.(2023)Lanham, Chen, Radhakrishnan, Steiner, Denison, Hernandez, Li, Durmus, Hubinger, Kernion, Lukosiute, Nguyen, Cheng, Joseph, Schiefer, Rausch, Larson, McCandlish, Kundu, Kadavath, Yang, Henighan, Maxwell, Telleen{-}Lawton, Hume, Hatfield{-}Dodds, Kaplan, Brauner, Bowman, and Perez]{DBLP:journals/corr/abs-2307-13702}
Tamera Lanham, Anna Chen, Ansh Radhakrishnan, Benoit Steiner, Carson Denison, Danny Hernandez, Dustin Li, Esin Durmus, Evan Hubinger, Jackson Kernion, Kamile Lukosiute, Karina Nguyen, Newton Cheng, Nicholas Joseph, Nicholas Schiefer, Oliver Rausch, Robin Larson, Sam McCandlish, Sandipan Kundu, Saurav Kadavath, Shannon Yang, Thomas Henighan, Timothy Maxwell, Timothy Telleen{-}Lawton, Tristan Hume, Zac Hatfield{-}Dodds, Jared Kaplan, Jan Brauner, Samuel~R. Bowman, and Ethan Perez.
\newblock Measuring faithfulness in chain-of-thought reasoning.
\newblock \emph{CoRR}, abs/2307.13702, 2023.
\newblock \doi{10.48550/arXiv.2307.13702}.
\newblock URL \url{https://doi.org/10.48550/arXiv.2307.13702}.

\bibitem[Lee \& Luo(2018)Lee and Luo]{lee2018strong}
Joohyung Lee and Man Luo.
\newblock Strong equivalence for lpmln programs.
\newblock In \emph{35th International Conference on Logic Programming (ICLP 2019)}, 2018.

\bibitem[Lee \& Wang(2016)Lee and Wang]{lee2016weighted}
Joohyung Lee and Yi~Wang.
\newblock Weighted rules under the stable model semantics.
\newblock In \emph{KRR}, 2016.

\bibitem[Lifschitz(2019)]{lifschitz2019answer}
Vladimir Lifschitz.
\newblock \emph{Answer set programming}.
\newblock Springer Berlin, 2019.

\bibitem[Liu et~al.(2022)Liu, Swayamdipta, Smith, and Choi]{liu2022wanli}
Alisa Liu, Swabha Swayamdipta, Noah~A Smith, and Yejin Choi.
\newblock Wanli: Worker and ai collaboration for natural language inference dataset creation.
\newblock \emph{arXiv preprint arXiv:2201.05955}, 2022.

\bibitem[Liu et~al.(2023{\natexlab{a}})Liu, Liu, Cui, Teng, Duan, Zhou, and Zhang]{liu2023logiQA}
Hanmeng Liu, Jian Liu, Leyang Cui, Zhiyang Teng, Nan Duan, Ming Zhou, and Yue Zhang.
\newblock Logiqa 2.0—an improved dataset for logical reasoning in natural language understanding.
\newblock \emph{IEEE/ACM Transactions on Audio, Speech, and Language Processing}, 31:\penalty0 2947--2962, 2023{\natexlab{a}}.
\newblock \doi{10.1109/TASLP.2023.3293046}.

\bibitem[Liu et~al.(2023{\natexlab{b}})Liu, Ning, Teng, Liu, Zhou, and Zhang]{liu2023evaluating}
Hanmeng Liu, Ruoxi Ning, Zhiyang Teng, Jian Liu, Qiji Zhou, and Yue Zhang.
\newblock Evaluating the logical reasoning ability of chatgpt and gpt-4.
\newblock \emph{arXiv preprint arXiv:2304.03439}, 2023{\natexlab{b}}.

\bibitem[Liu et~al.(2023{\natexlab{c}})Liu, Teng, Cui, Zhang, Zhou, and Zhang]{liu2023logicot}
Hanmeng Liu, Zhiyang Teng, Leyang Cui, Chaoli Zhang, Qiji Zhou, and Yue Zhang.
\newblock Logicot: Logical chain-of-thought instruction-tuning data collection with gpt-4.
\newblock \emph{arXiv preprint arXiv:2305.12147}, 2023{\natexlab{c}}.

\bibitem[Liu et~al.(2020)Liu, Cui, Liu, Huang, Wang, and Zhang]{liu2020logiQA}
Jian Liu, Leyang Cui, Hanmeng Liu, Dandan Huang, Yile Wang, and Yue Zhang.
\newblock Logiqa: A challenge dataset for machine reading comprehension with logical reasoning.
\newblock In \emph{IJCAI}, 2020.
\newblock \doi{10.24963/ijcai.2020/501}.
\newblock URL \url{https://doi.org/10.24963/ijcai.2020/501}.

\bibitem[Liu et~al.(2021)Liu, Cui, Liu, Huang, Wang, and Zhang]{liu2021logiqa}
Jian Liu, Leyang Cui, Hanmeng Liu, Dandan Huang, Yile Wang, and Yue Zhang.
\newblock Logiqa: a challenge dataset for machine reading comprehension with logical reasoning.
\newblock In \emph{Proceedings of the Twenty-Ninth International Conference on International Joint Conferences on Artificial Intelligence}, pp.\  3622--3628, 2021.

\bibitem[Liu et~al.(2019)Liu, Ott, Goyal, and et~al.]{liu2019roberta}
Yinhan Liu, Myle Ott, Naman Goyal, and et~al.
\newblock Roberta: A robustly optimized bert pretraining approach.
\newblock \emph{arXiv:1907.11692}, 2019.

\bibitem[Lourie et~al.(2021)Lourie, Bras, Bhagavatula, and Choi]{lourie2021unicorn}
Nicholas Lourie, Ronan~Le Bras, Chandra Bhagavatula, and Yejin Choi.
\newblock Unicorn on rainbow: A universal commonsense reasoning model on a new multitask benchmark.
\newblock \emph{arXiv preprint arXiv:2103.13009}, 2021.

\bibitem[Lu et~al.(2023)Lu, Peng, Cheng, Galley, Chang, Wu, Zhu, and Gao]{lu2023chameleon}
Pan Lu, Baolin Peng, Hao Cheng, Michel Galley, Kai-Wei Chang, Ying~Nian Wu, Song-Chun Zhu, and Jianfeng Gao.
\newblock Chameleon: Plug-and-play compositional reasoning with large language models.
\newblock \emph{arXiv preprint arXiv:2304.09842}, 2023.

\bibitem[Luo et~al.(2022)Luo, Hashimoto, Yavuz, Liu, Baral, and Zhou]{luo2022choose}
Man Luo, Kazuma Hashimoto, Semih Yavuz, Zhiwei Liu, Chitta Baral, and Yingbo Zhou.
\newblock Choose your qa model wisely: A systematic study of generative and extractive readers for question answering.
\newblock \emph{Spa-NLP 2022}, pp.\ ~7, 2022.

\bibitem[McCarthy(1989)]{mccarthy1989artificial}
John McCarthy.
\newblock Artificial intelligence, logic and formalizing common sense.
\newblock In \emph{Philosophical logic and artificial intelligence}. Springer, 1989.

\bibitem[Mirzaee et~al.(2021)Mirzaee, Rajaby~Faghihi, Ning, and Kordjamshidi]{mirzaee2021spartqa}
Roshanak Mirzaee, Hossein Rajaby~Faghihi, Qiang Ning, and Parisa Kordjamshidi.
\newblock {SPARTQA}: A textual question answering benchmark for spatial reasoning.
\newblock In Kristina Toutanova, Anna Rumshisky, Luke Zettlemoyer, Dilek Hakkani-Tur, Iz~Beltagy, Steven Bethard, Ryan Cotterell, Tanmoy Chakraborty, and Yichao Zhou (eds.), \emph{Proceedings of the 2021 Conference of the North American Chapter of the Association for Computational Linguistics: Human Language Technologies}, pp.\  4582--4598, Online, June 2021. Association for Computational Linguistics.
\newblock \doi{10.18653/v1/2021.naacl-main.364}.
\newblock URL \url{https://aclanthology.org/2021.naacl-main.364}.

\bibitem[Mishra et~al.(2022{\natexlab{a}})Mishra, Khashabi, Baral, and Hajishirzi]{mishra2022cross}
Swaroop Mishra, Daniel Khashabi, Chitta Baral, and Hannaneh Hajishirzi.
\newblock Cross-task generalization via natural language crowdsourcing instructions.
\newblock In \emph{Proceedings of the 60th Annual Meeting of the Association for Computational Linguistics (Volume 1: Long Papers)}, pp.\  3470--3487, 2022{\natexlab{a}}.

\bibitem[Mishra et~al.(2022{\natexlab{b}})Mishra, Mitra, Varshney, Sachdeva, Clark, Baral, and Kalyan]{mishra2022numglue}
Swaroop Mishra, Arindam Mitra, Neeraj Varshney, Bhavdeep Sachdeva, Peter Clark, Chitta Baral, and Ashwin Kalyan.
\newblock Numglue: A suite of fundamental yet challenging mathematical reasoning tasks.
\newblock In \emph{Proceedings of the 60th Annual Meeting of the Association for Computational Linguistics (Volume 1: Long Papers)}, pp.\  3505--3523, 2022{\natexlab{b}}.

\bibitem[Namgyu~Ho(2022)]{NamgyuHo2022}
Se-Young~Yun Namgyu~Ho, Laura~Schmid.
\newblock Large language models are reasoning teachers.
\newblock \emph{arXiv preprint arXiv:2212.10071}, 2022.

\bibitem[Nie et~al.(2020)Nie, Williams, Dinan, Bansal, Weston, and Kiela]{nie2020adversarial}
Yixin Nie, Adina Williams, Emily Dinan, Mohit Bansal, Jason Weston, and Douwe Kiela.
\newblock Adversarial nli: A new benchmark for natural language understanding.
\newblock In \emph{Proceedings of the 58th Annual Meeting of the Association for Computational Linguistics}, pp.\  4885--4901, 2020.

\bibitem[Niven \& Kao(2019)Niven and Kao]{niven2019probing}
Timothy Niven and Hung-Yu Kao.
\newblock Probing neural network comprehension of natural language arguments.
\newblock In \emph{ACL}, 2019.

\bibitem[Olausson* et~al.(2023)Olausson*, Gu*, Lipkin*, Zhang*, Solar-Lezama, Tenenbaum, and Levy]{OGLZ_LINC_2023}
Theo~X. Olausson*, Alex Gu*, Ben Lipkin*, Cedegao~E. Zhang*, Armando Solar-Lezama, Joshua~B. Tenenbaum, and Roger~P. Levy.
\newblock Linc: A neuro-symbolic approach for logical reasoning by combining language models with first-order logic provers.
\newblock 2023.

\bibitem[Pan et~al.(2023)Pan, Albalak, Wang, and Wang]{pan2023logic}
Liangming Pan, Alon Albalak, Xinyi Wang, and William~Yang Wang.
\newblock Logic-lm: Empowering large language models with symbolic solvers for faithful logical reasoning.
\newblock \emph{arXiv preprint arXiv:2305.12295}, 2023.

\bibitem[Paranjape et~al.(2023)Paranjape, Lundberg, Singh, Hajishirzi, Zettlemoyer, and Ribeiro]{paranjape2023art}
Bhargavi Paranjape, Scott Lundberg, Sameer Singh, Hannaneh Hajishirzi, Luke Zettlemoyer, and Marco~Tulio Ribeiro.
\newblock Art: Automatic multi-step reasoning and tool-use for large language models.
\newblock \emph{arXiv preprint arXiv:2303.09014}, 2023.

\bibitem[Parmar et~al.(2022)Parmar, Mishra, Purohit, Luo, Mohammad, and Baral]{parmar2022boxbart}
Mihir Parmar, Swaroop Mishra, Mirali Purohit, Man Luo, Murad Mohammad, and Chitta Baral.
\newblock In-boxbart: Get instructions into biomedical multi-task learning.
\newblock In \emph{Findings of the Association for Computational Linguistics: NAACL 2022}, pp.\  112--128, 2022.

\bibitem[Paul(1993)]{paul1993approaches}
Gabriele Paul.
\newblock Approaches to abductive reasoning: an overview.
\newblock \emph{Artificial intelligence review}, 7\penalty0 (2):\penalty0 109--152, 1993.

\bibitem[Radford et~al.()Radford, Wu, Child, Luan, Amodei, Sutskever, et~al.]{radford2019language}
Alec Radford, Jeffrey Wu, Rewon Child, David Luan, Dario Amodei, Ilya Sutskever, et~al.
\newblock Language models are unsupervised multitask learners.

\bibitem[Richardson \& Sabharwal(2021)Richardson and Sabharwal]{richardson2021nlsat}
Kyle Richardson and Ashish Sabharwal.
\newblock Pushing the limits of rule reasoning in transformers through natural language satisfiability.
\newblock \emph{arXiv:2112.09054}, 2021.

\bibitem[Richardson \& Domingos(2006)Richardson and Domingos]{mlnrichardson2006}
Matthew Richardson and Pedro Domingos.
\newblock Markov logic networks.
\newblock \emph{Mach. Learn.}, 62\penalty0 (1–2):\penalty0 107–136, feb 2006.
\newblock ISSN 0885-6125.
\newblock \doi{10.1007/s10994-006-5833-1}.
\newblock URL \url{https://doi.org/10.1007/s10994-006-5833-1}.

\bibitem[Saeed et~al.(2021)Saeed, Ahmadi, Nakov, and Papotti]{saeed2021rulebert}
Mohammed Saeed, Naser Ahmadi, Preslav Nakov, and Paolo Papotti.
\newblock {R}ule{BERT}: Teaching soft rules to pre-trained lms.
\newblock In \emph{EMNLP}, November 2021.
\newblock URL \url{https://aclanthology.org/2021.emnlp-main.110}.

\bibitem[Saha et~al.(2020)Saha, Ghosh, Srivastava, and Bansal]{saha2020prover}
Swarnadeep Saha, Sayan Ghosh, Shashank Srivastava, and Mohit Bansal.
\newblock Prover: Proof generation for interpretable reasoning over rules.
\newblock In \emph{EMNLP}, 2020.

\bibitem[Sanyal et~al.(2022)Sanyal, Xu, Wang, Yang, Pryzant, Yu, Zhu, and Ren]{sanyal2022apollo}
Soumya Sanyal, Yichong Xu, Shuohang Wang, Ziyi Yang, Reid Pryzant, Wenhao Yu, Chenguang Zhu, and Xiang Ren.
\newblock Apollo: A simple approach for adaptive pretraining of language models for logical reasoning.
\newblock \emph{arXiv preprint arXiv:2212.09282}, 2022.

\bibitem[Saparov \& He(2023{\natexlab{a}})Saparov and He]{SaparovHe2023}
Abulhair Saparov and He~He.
\newblock Language models are greedy reasoners: A systematic formal analysis of chain-of-thought.
\newblock In \emph{The Eleventh International Conference on Learning Representations}, 2023{\natexlab{a}}.
\newblock URL \url{https://openreview.net/forum?id=qFVVBzXxR2V}.

\bibitem[Saparov \& He(2023{\natexlab{b}})Saparov and He]{abulhair}
Abulhair Saparov and He~He.
\newblock Language models are greedy reasoners: A systematic formal analysis of chain-of-thought.
\newblock In \emph{The Eleventh International Conference on Learning Representations}, 2023{\natexlab{b}}.
\newblock URL \url{https://openreview.net/forum?id=qFVVBzXxR2V}.

\bibitem[Saparov et~al.(2023)Saparov, Pang, Padmakumar, Joshi, Kazemi, Kim, and He]{saparov2023testing}
Abulhair Saparov, Richard~Yuanzhe Pang, Vishakh Padmakumar, Nitish Joshi, Seyed~Mehran Kazemi, Najoung Kim, and He~He.
\newblock Testing the general deductive reasoning capacity of large language models using ood examples.
\newblock \emph{arXiv preprint arXiv:2305.15269}, 2023.

\bibitem[Sauce \& Matzel(2017)Sauce and Matzel]{sauce2017inductive}
Bruno Sauce and Louis~D Matzel.
\newblock Inductive reasoning.
\newblock \emph{Encyclopedia of animal cognition and behavior}, 6:\penalty0 1--8, 2017.

\bibitem[Schick et~al.(2023)Schick, Dwivedi-Yu, Dess{\`\i}, Raileanu, Lomeli, Zettlemoyer, Cancedda, and Scialom]{schick2023toolformer}
Timo Schick, Jane Dwivedi-Yu, Roberto Dess{\`\i}, Roberta Raileanu, Maria Lomeli, Luke Zettlemoyer, Nicola Cancedda, and Thomas Scialom.
\newblock Toolformer: Language models can teach themselves to use tools.
\newblock \emph{arXiv preprint arXiv:2302.04761}, 2023.

\bibitem[Sinha et~al.(2019)Sinha, Sodhani, Dong, and et~al.]{sinha2019clutrr}
Koustuv Sinha, Shagun Sodhani, Jin Dong, and et~al.
\newblock Clutrr: A diagnostic benchmark for inductive reasoning from text.
\newblock In \emph{EMNLP}, 2019.

\bibitem[Speer et~al.(2017)Speer, Chin, and Havasi]{speer2017conceptnet}
Robyn Speer, Joshua Chin, and Catherine Havasi.
\newblock Conceptnet 5.5: An open multilingual graph of general knowledge.
\newblock In \emph{Proceedings of the AAAI conference on artificial intelligence}, volume~31, 2017.

\bibitem[Srivastava et~al.(2022)Srivastava, Rastogi, Rao, Shoeb, Abid, Fisch, Brown, Santoro, Gupta, Garriga-Alonso, et~al.]{srivastava2022beyond}
Aarohi Srivastava, Abhinav Rastogi, Abhishek Rao, Abu Awal~Md Shoeb, Abubakar Abid, Adam Fisch, Adam~R Brown, Adam Santoro, Aditya Gupta, Adri{\`a} Garriga-Alonso, et~al.
\newblock Beyond the imitation game: Quantifying and extrapolating the capabilities of language models.
\newblock \emph{arXiv preprint arXiv:2206.04615}, 2022.

\bibitem[Srivastava et~al.(2023)Srivastava, Rastogi, Rao, Shoeb, Abid, Fisch, Brown, Santoro, Gupta, Garriga-Alonso, et~al.]{srivastava2023beyond}
Aarohi Srivastava, Abhinav Rastogi, Abhishek Rao, Abu Awal~Md Shoeb, Abubakar Abid, Adam Fisch, Adam~R Brown, Adam Santoro, Aditya Gupta, Adri{\`a} Garriga-Alonso, et~al.
\newblock Beyond the imitation game: Quantifying and extrapolating the capabilities of language models.
\newblock \emph{Transactions on Machine Learning Research}, 2023.

\bibitem[Sumers et~al.(2023)Sumers, Yao, Narasimhan, and Griffiths]{sumers2023cognitive}
Theodore Sumers, Shunyu Yao, Karthik Narasimhan, and Thomas~L Griffiths.
\newblock Cognitive architectures for language agents.
\newblock \emph{arXiv preprint arXiv:2309.02427}, 2023.

\bibitem[Tafjord et~al.(2021)Tafjord, Dalvi, and Clark]{tafjord2021proofwriter}
Oyvind Tafjord, Bhavana Dalvi, and Peter Clark.
\newblock {P}roof{W}riter: Generating implications, proofs, and abductive statements over natural language.
\newblock In \emph{Findings-ACL-IJCNLP}, August 2021.
\newblock \doi{10.18653/v1/2021.findings-acl.317}.
\newblock URL \url{https://aclanthology.org/2021.findings-acl.317}.

\bibitem[Tandon et~al.(2019)Tandon, Dalvi, Sakaguchi, Clark, and Bosselut]{tandon2019wiqa}
Niket Tandon, Bhavana Dalvi, Keisuke Sakaguchi, Peter Clark, and Antoine Bosselut.
\newblock Wiqa: A dataset for “what if...” reasoning over procedural text.
\newblock In \emph{Proceedings of the 2019 Conference on Empirical Methods in Natural Language Processing and the 9th International Joint Conference on Natural Language Processing (EMNLP-IJCNLP)}, pp.\  6076--6085, 2019.

\bibitem[Teng et~al.(2023)Teng, Ning, Liu, Zhou, Zhang, et~al.]{teng2023glore}
Zhiyang Teng, Ruoxi Ning, Jian Liu, Qiji Zhou, Yue Zhang, et~al.
\newblock Glore: Evaluating logical reasoning of large language models.
\newblock \emph{arXiv preprint arXiv:2310.09107}, 2023.

\bibitem[Tian et~al.(2021{\natexlab{a}})Tian, Li, Chen, Xiao, He, and Jin]{tian2021diagnosing}
Jidong Tian, Yitian Li, Wenqing Chen, Liqiang Xiao, Hao He, and Yaohui Jin.
\newblock Diagnosing the first-order logical reasoning ability through logicnli.
\newblock In \emph{Proceedings of the 2021 Conference on Empirical Methods in Natural Language Processing}, pp.\  3738--3747, 2021{\natexlab{a}}.

\bibitem[Tian et~al.(2021{\natexlab{b}})Tian, Li, Chen, Xiao, He, and Jin]{tian2021logicnli}
Jidong Tian, Yitian Li, Wenqing Chen, Liqiang Xiao, Hao He, and Yaohui Jin.
\newblock Diagnosing the first-order logical reasoning ability through {L}ogic{NLI}.
\newblock In \emph{EMNLP}, November 2021{\natexlab{b}}.
\newblock URL \url{https://aclanthology.org/2021.emnlp-main.303}.

\bibitem[Varshney et~al.(2023)Varshney, Parmar, Patel, Handa, Sarkar, Luo, and Baral]{varshney2023can}
Neeraj Varshney, Mihir Parmar, Nisarg Patel, Divij Handa, Sayantan Sarkar, Man Luo, and Chitta Baral.
\newblock Can nlp models correctly reason over contexts that break the common assumptions?
\newblock \emph{arXiv preprint arXiv:2305.12096}, 2023.

\bibitem[Vaswani et~al.(2017)Vaswani, Shazeer, Parmar, and et~al.]{Vaswani2017transformer}
Ashish Vaswani, Noam Shazeer, Niki Parmar, and et~al.
\newblock Attention is all you need.
\newblock In \emph{NeurIPS}, volume~30. Curran Associates, Inc., 2017.
\newblock URL \url{https://proceedings.neurips.cc/paper/2017/file/3f5ee243547dee91fbd053c1c4a845aa-Paper.pdf}.

\bibitem[Wang et~al.(2019)Wang, Pruksachatkun, Nangia, Singh, Michael, Hill, Levy, and Bowman]{wang2019superglue}
Alex Wang, Yada Pruksachatkun, Nikita Nangia, Amanpreet Singh, Julian Michael, Felix Hill, Omer Levy, and Samuel Bowman.
\newblock Superglue: A stickier benchmark for general-purpose language understanding systems.
\newblock \emph{Advances in neural information processing systems}, 32, 2019.

\bibitem[Wei et~al.(2022)Wei, Wang, Schuurmans, Bosma, Xia, Chi, Le, Zhou, et~al.]{wei2022chain}
Jason Wei, Xuezhi Wang, Dale Schuurmans, Maarten Bosma, Fei Xia, Ed~Chi, Quoc~V Le, Denny Zhou, et~al.
\newblock Chain-of-thought prompting elicits reasoning in large language models.
\newblock \emph{Advances in Neural Information Processing Systems}, 35:\penalty0 24824--24837, 2022.

\bibitem[Weston et~al.(2015{\natexlab{a}})Weston, Bordes, Chopra, Rush, Van~Merri{\"e}nboer, Joulin, and Mikolov]{weston2015towards}
Jason Weston, Antoine Bordes, Sumit Chopra, Alexander~M Rush, Bart Van~Merri{\"e}nboer, Armand Joulin, and Tomas Mikolov.
\newblock Towards ai-complete question answering: A set of prerequisite toy tasks.
\newblock \emph{arXiv preprint arXiv:1502.05698}, 2015{\natexlab{a}}.

\bibitem[Weston et~al.(2015{\natexlab{b}})Weston, Bordes, and et~al.]{weston2015babi}
Jason Weston, Antoine Bordes, and et~al.
\newblock Towards ai-complete question answering: A set of prerequisite toy tasks.
\newblock \emph{arXiv:1502.05698}, 2015{\natexlab{b}}.

\bibitem[Wu et~al.(2023)Wu, Qiu, Ross, Aky{\"{u}}rek, Chen, Wang, Kim, Andreas, and Kim]{DBLP:journals/corr/abs-2307-02477}
Zhaofeng Wu, Linlu Qiu, Alexis Ross, Ekin Aky{\"{u}}rek, Boyuan Chen, Bailin Wang, Najoung Kim, Jacob Andreas, and Yoon Kim.
\newblock Reasoning or reciting? exploring the capabilities and limitations of language models through counterfactual tasks.
\newblock \emph{CoRR}, abs/2307.02477, 2023.
\newblock \doi{10.48550/arXiv.2307.02477}.
\newblock URL \url{https://doi.org/10.48550/arXiv.2307.02477}.

\bibitem[Yang et~al.(2019)Yang, Dai, Yang, and Carbonell]{Yang2019xlnet}
Zhilin Yang, Zihang Dai, Yiming Yang, and et~al. Carbonell.
\newblock Xlnet: Generalized autoregressive pretraining for language understanding.
\newblock In \emph{NeurIPS}, volume~32. Curran Associates, Inc., 2019.
\newblock URL \url{https://proceedings.neurips.cc/paper/2019/file/dc6a7e655d7e5840e66733e9ee67cc69-Paper.pdf}.

\bibitem[Yu et~al.(2019)Yu, Jiang, Dong, and Feng]{yu2019reclor}
Weihao Yu, Zihang Jiang, Yanfei Dong, and Jiashi Feng.
\newblock Reclor: A reading comprehension dataset requiring logical reasoning.
\newblock In \emph{ICLR}, 2019.

\bibitem[Yu et~al.(2020)Yu, Jiang, Dong, and Feng]{yureclor}
Weihao Yu, Zihang Jiang, Yanfei Dong, and Jiashi Feng.
\newblock Reclor: A reading comprehension dataset requiring logical reasoning.
\newblock In \emph{International Conference on Learning Representations}, 2020.

\bibitem[Zhang et~al.(2022)Zhang, Li, Meng, Chang, and Broeck]{zhang2022paradox}
Honghua Zhang, Liunian~Harold Li, Tao Meng, Kai-Wei Chang, and Guy Van~den Broeck.
\newblock On the paradox of learning to reason from data.
\newblock \emph{arXiv preprint arXiv:2205.11502}, 2022.

\end{thebibliography}
\bibliographystyle{colm2024_conference}

\appendix
\newpage
\appendix 

\section{Dataset Licence}
\label{apd:license}
All datasets included in LogiGLUE benchmark are covered under licenses: Apache-2.0, MIT License, Attribution-NonCommercial 4.0 International (CC BY-NC 4.0), and BSD License. Each dataset license is given in Table~\ref{tab:license}.  These licenses generally permit the reproduction of the datasets for non-commercial purposes. LogiGLUE will be public available under the license of Attribution-NonCommercial 4.0 International license.

\begin{table}[t]
\centering
 \resizebox{0.7\linewidth}{!}{
\begin{tabular}{l|l}
    \toprule
     {\bf Dataset}  &{\bf License} \\
     \toprule
     Abductive NLI & Apache-2.0 license \\
    CLUTTR &  Attribution-NonCommercial 4.0 International \\
    Abduction Rules &  MIT License \\
    Adversarial NLI &  Attribution-NonCommercial 4.0 International \\
    LogiNLI &   Creative Commons Attribution 4.0 International License \\
    proofWriter & Creative Commons Attribution 4.0 International License \\ 
    RuleBERT &  Creative Commons Attribution 4.0 International License \\
    Folio & CC-BY-SA-4.0 license \\
    LogiQA &  CC BY-NC-SA 4.0 DEED \\
    baBi &  BSD License \\
    Electric-bird &  Apache-2.0 license  \\
    Winologic &  Creative Commons Attribution 4.0 International License \\
    Wanli & Creative Commons Attribution 4.0 International License \\
    BigBench &  Apache-2.0 license \\
    ProntoQA &  Apache-2.0 license\\
    \bottomrule
    \end{tabular}
    }
\caption{Dataset License.}
\label{tab:license}
\end{table}

\section{Compare with Existing Logical 
Reasoning Benchmark}
\label{apd:other_benchmark}
GLoRe~\citep{teng2023glore} offers only an evaluation dataset, lacking associated training data. In contrast, our dataset includes training sets formatted in a distinct and innovative style. Unlike LogiTorch~\citep{helwe2022logitorch}, which does not specify the reasoning types for its datasets, LogiGLUE categorizes each dataset into either deductive/inductive/abductive reasoning, also encompassing more datasets compared to LogiTorch. While LogiBench~\citep{anonymous2023logicbench} is an extensive dataset, it is limited to deductive reasoning and solely consists of synthetically generated data. Furthermore, CROW~\citep{ismayilzada-etal-2023-crow} primarily concentrates on commonsense reasoning and is exclusively designed for evaluative purposes.

\section{Additional Experiments Results}

\subsection{Detailed Fine-tuning Results} 
In Table~\ref{tab:indomain_result}, we provide the detailed performance of each model that trained with single-task, multi-task and multi-task with single task training methods on each in-domain datasets. 
\begin{table*}[t]
\centering
 \resizebox{\linewidth}{!}{
\begin{tabular}{l|c|c|c}
    \toprule
     {\bf Dataset}  &  {\bf Single-Task (Flan-T5-large)} &  {\bf Multi-Task (LogiT5)} & {\bf Single Task (LogiT5)} \\
     \toprule
        $\alpha$ARCT~\citeyear{niven2019probing} & 72.31 & \textbf{77.22} & 76.74 \\
        $\alpha$NLI~\citeyear{bhagavatulaabductive}  & 78.26  & 76.37 & \textbf{78.46} \\
        CLUTTR-Robust~\citeyear{sinha2019clutrr}  & \textbf{97.22} & 96.53& \textbf{97.22}  \\   AbductionRule-Animal ~\citeyear{ignatiev2019abduction} & \textbf{100} & \textbf{100}& \textbf{100} \\ ANLI~\citeyear{nie2020adversarial}  & \textbf{61.16} & 59.53& 60.53\\  
        LogiQA~\citeyear{liu2021logiqa}  & 37.94 & 38.56 & \textbf{39.94}\\
        LogicNLI~\citeyear{tian2021diagnosing}   & 82.60 & 88.40 & \textbf{88.65}\\
        ProofWriter~\citeyear{tafjord2021proofwriter}   & 99.42 & 98.85 & \textbf{99.55} \\
        Rulebert-Union~\citeyear{saeed2021rulebert}   & 99.69 & 99.36 & \textbf{99.70}\\
        FOLIO~\citeyear{han2022folio} & 66.66 & \textbf{74.02}&  72.06 \\
        \midrule
        Average &79.52 &80.88 & \textbf{81.28}\\
    \bottomrule
    \end{tabular}
    }
\caption{Three training strategies for models and the performance on In-domain Testing Dataset.}
\label{tab:indomain_result}
\end{table*}

\subsection{CoT Distillation Fine-tuning }
With the CoT fine-tuning, we observed that the fine-tuning takes longer time and a larger learning rate in the beginning is helpful. Thus, instead of using 1e-4 as the learning rate, we trained the model with 40 epochs and a learning rate of 3e-4 we use. We do see that the model performance increase when the number of epochs increase. 

\subsection{Zero-shot Evaluation with CoT Prompting Format}
\label{apd:zero-shot-prompt}

In the zero-shot evaluation of Llama,  GPT-3 and GPT-4, we add a ``Only based on the context, let’s think step by step.'' sentence in the end to prompt models to generate reasoning. 
Take Winologic dataset as an example, the input to the model is following:

Context: I can't cut that tree down with that axe; it is too thick.
Question: The it is more likely to refer to the tree than the axe because when the person is using the axe to cut down the tree, if the axe is too thick, the person might not be successful.
Options: 
 - False
 - True
Please select the correct option. \textbf{Only based on the context, let’s think step by step. }

\subsection{Full Evaluation On Out-of-Domain Datasets}
\label{apd:ood}
In Table~\ref{tab:outdomain_result_apd}, we present comprehensive results for all out-of-domain datasets. Due to the financial constraints, we've excluded the complete GPT-4 and GPT-3.5-turbo outcomes on the full set. However, our primary results of GPT-4 and  GPT-3.5-turbo tested on a randomly selected 20\% of the test set are presented in the main paper, which can be taken as an approximation of full-set performance. 
It is worth to mentioning that GPT-3.5 and LLama-2 produce free-form text, making evaluation challenging, while Flan-T5's structured output, trained on instruction fine-tuned data, simplifies this process. LLama-2 initially showed poor results, but upon using ConceptNET~\cite{speer2017conceptnet} to account for synonyms, its performance improved. The babi dataset revealed LLama's tendency to rely on pre-trained knowledge, ignoring input text, a phenomenon also noted in~\citet{varshney2023can}.
Our analysis also includes a comparison of reasoning processes with and without Chain of Thought (CoT) prompting. 
While CoT offers a modest advantage, the impact is not markedly significant. We attribute this to LLaMA's inherent tendency to generate reasoning chains, even without explicit step-by-step reasoning prompts.

\begin{table*}[!htb]
\centering
 \resizebox{\linewidth}{!}{
\begin{tabular}{l|c|c|c|c}
    \toprule
     {\bf Dataset}  &  {\bf Flan-T5-large} &  {\bf LogiT5}  &  {\bf LLAMA-2-7b-chat}  &  {\bf LLAMA-2-7b-chat-CoT} \\
     \toprule
        bAbi(induction)~\citeyear{weston2015towards}&\textbf{59.44} & 13.12  &  33.46 & 29.04 \\
        bAbi(deduction)~\citeyear{weston2015towards} & 40.18  & 65.84 &  58.76& \textbf{64.84} \\
        CLUTTR-Systematic~\citeyear{sinha2019clutrr}   &  15.43 & \textbf{100 }&  38.82& 30.62\\
        AbductionRule-person ~\citeyear{ignatiev2019abduction}  &  00.00 & \textbf{95.97} &  31.31 & 54.91\\        
        ReClor~\citeyear{yureclor}   &  40.20 & \textbf{47.00} &  39.80 & 40.20\\        
        Bird-Electricity~\citeyear{tafjord2021proofwriter} &  41.12 & \textbf{65.29} &  50.00& 38.78\\
        NatlLang~\citeyear{tafjord2021proofwriter}  &  56.03 & \textbf{70.24} &  49.52& 50.29\\
        Winologic~\citeyear{he2021winologic}  &  \textbf{68.51} & 61.39 &  48.93 & 49.28\\
        WaNLI~\citeyear{liu2022wanli} &  49.88 & \textbf{62.22} &  32.80 & 32.32 \\
        Rulebert-Union~\citeyear{saeed2021rulebert} &  24.92 & 62.34 &  {50.02} & 52.26\\
        BigBench(logical-deduction)~\citeyear{srivastava2022beyond} &  \textbf{46.31} & 30.92 &  22.38& 25.61\\
        BigBench(logical args)~\citeyear{srivastava2022beyond}  &  \textbf{43.75} & 40.62 &  40.62 & 31.25\\
        LogiQA 2.0 ~\citeyear{liu2023logiQA}  &  \textbf{52.66} & 23.13 & 50.74  & 50.15 \\
        PrOntoQA-OOD ~\citeyear{saparov2023testing}  &  6.50 & 33.50 & \textbf{60.20} &66.5 \\        
        \midrule
        Average &38.70 &\textbf{55.11}& 43.38& 44.00\\
    \bottomrule
    \end{tabular}
    }
\caption{Performance on Leave-out Testing Datasets.}
\label{tab:outdomain_result_apd}
\end{table*}

\end{document}